\documentclass[sigconf]{acmart}

\usepackage{algorithm}
\usepackage{algorithmicx}
\usepackage{amsmath}
\usepackage{graphics}
\usepackage{algpseudocode}
\usepackage{multicol}
\usepackage{multirow}
\usepackage{subfigure}
\usepackage{epsfig}
\usepackage{enumitem}
\usepackage{mathrsfs}
\usepackage{bm}
\usepackage{graphicx}
\usepackage{epstopdf}
\usepackage{float}
\usepackage{verbatim}
\usepackage{array}
\usepackage{balance}
\usepackage{arydshln}
\usepackage{bbding}

\AtBeginDocument{%
  \providecommand\BibTeX{{%
    \normalfont B\kern-0.5em{\scshape i\kern-0.25em b}\kern-0.8em\TeX}}}

\copyrightyear{2022}
\acmYear{2022}
\setcopyright{acmcopyright}
\acmConference[MM '22]{Proceedings of the 30th ACM International Conference on Multimedia}{October 10--14, 2022}{Lisbon, Portugal}
\acmBooktitle{Proceedings of the 30th ACM International Conference on Multimedia (MM '22), October 10--14, 2022, Lisbon, Portugal}

\begin{document}

\fancyhead{}

\title{Three-Stream Joint Network  for Zero-Shot \\
Sketch-Based Image Retrieval}
\author{Yu-Wei Zhan$^{1}$, Xin Luo$^{1}$, Yongxin Wang$^{2}$, Zhen-Duo Chen$^{1}$, Xin-Shun Xu$^{1}$}
\author{$^{1}$School of Software, Shandong University, Jinan 250101, China}
\author{$^{2}$School of Computer Science and Technology, Shandong Jianzhu University, Jinan 250101, China}
\author{\ }

\renewcommand{\shortauthors}{Trovato and Tobin, et al.}

\begin{abstract}

The Zero-Shot Sketch-based Image Retrieval (ZS-SBIR) is a challenging task because of the large domain gap between sketches and natural images as well as the semantic inconsistency between seen and unseen categories. Previous literature bridges seen and unseen categories by semantic embedding, which requires prior knowledge of the exact class names and additional extraction efforts. And most works reduce domain gap by mapping sketches and natural images into a common high-level space using constructed sketch-image pairs, which ignore the unpaired information between images and sketches. To address these issues, in this paper, we propose a novel Three-Stream Joint Training Network (3JOIN) for the ZS-SBIR task. To narrow the domain differences between sketches and images, we extract edge maps for natural images and treat them as a bridge between images and sketches, which have similar content to images and similar style to sketches. For exploiting a sufficient combination of sketches, natural images, and edge maps, a novel three-stream joint training network is proposed.
In addition, we use a teacher network to extract the implicit semantics of the samples without the aid of other semantics and transfer the learned knowledge to unseen classes. Extensive experiments conducted on two real-world datasets demonstrate the superiority of our proposed method.

\end{abstract}

\fancyhead{}

\maketitle
\section{Introduction}\label{introduction}

Given a hand-drawn sketch as a query and a large database of images as a gallery, Sketch-Based Image Retrieval (SBIR) aims at finding correlated images from the gallery, i.e., those with similar visual content or the same object class as the query. Compared with traditional text-image cross-modal retrieval, SBIR may become the preferred retrieval method for users when it is difficult to provide text descriptions but easy to sketch the desired content. Due to the rapid emergence and iteration of smart and touch screen devices, SBIR has attracted widespread attention with potential applications in e-commerce, forensics, and other fields. Existing SBIR works, trained on large-scale labeled datasets, have achieved amazing performance. However, SBIR has been facing an enormous challenge. With the explosive growth of multimedia data and newly emerging concepts on the Internet, it becomes impractical to include images of all categories to the training data. Therefore, the researchers situate the SIBR task in the condition of ZS-SBIR, which assumes that query sketches and database images in the target domain are unseen in the training phase.

\begin{figure}
\begin{minipage}{0.245\linewidth}\centering
\centerline{\includegraphics[height=7cm]{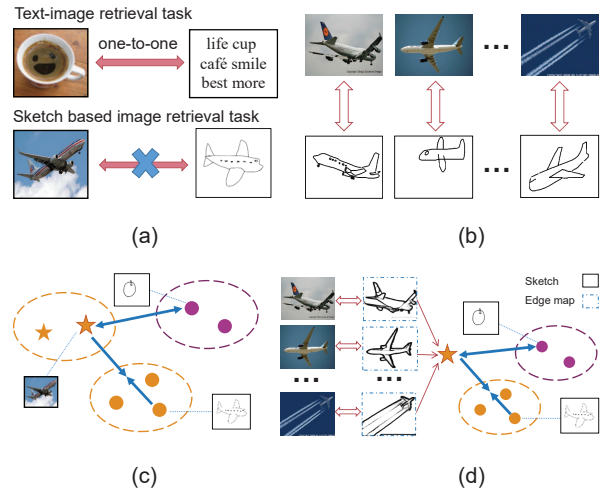}}
\end{minipage} \vspace{-0.3cm}
\caption{An illustration of our proposed method. (a) Differences
between text-image retrieval task and SBIR task. (b)
Narrow the domain gap by constructing image-sketch pairs.
(c) Narrow the domain gap by constructing triples. (d) Our
proposed method utilizes edge maps as a bridge between natural
images and sketches.}\label{motivation}\vspace{-0.3cm}
\end{figure}

ZS-SBIR, a cross-modal learning task under zero-shot learning, is non-trivial due to the following two challenges. The first crucial challenge is how to resolve the semantic inconsistency between seen and unseen classes in a simple and efficient fashion. The knowledge gap between seen and unseen classes makes the ZS-SBIR task even more intractable. ZS-SBIR requires the model capable of retrieving samples from the test set, but categories of the test set are unseen during the whole training time. In this case, it is necessary to establish a link between the training set and the test set, and transfer the knowledge learned from the training set to the test set. Several works have been proposed to address this issue.
Most of them bridge seen and unseen categories through semantic embeddings, i.e., extracting word vectors from NLP models or measuring word similarity through hierarchical models. However, such a strategy would require prior knowledge of the exact class name and additional extraction work, which would result in a burden of resources.

The second challenge is how to minimize the domain differences between modalities to enhance the representation power of retrieval codes for multi-modal data. There exists the well-known ``domain gap'' between sketches and natural images. Specifically, there are huge differences between sketches and natural images: 1) Different content. Natural images often contain complex backgrounds that may involve multiple objects, while sketches are abstract representations of a specific object. 2) Different styles. Natural images have rich colors and detailed textures, while sketches contain only the outline of an object. The existing methods implement ZS-SBIR task by exploring the matching relationship between sketch-image pairs. However, all of them fail to consider that the edge map extracted from the natural image can be used as a bridge between the natural image and the sketch. At present, there are many mature and simple edge map extraction algorithms that can convert natural images into the form of edge maps, which consist of a white background and black lines that represents the main outline of the object. All of these extracted edge maps have the same content as natural images and a similar style to sketches, which bridge between the natural image and the sketch to minimize the domain gap. Exploiting the fully joint between sketch, natural image, and edge map provides a new approach to solve cross-domain problems.


More importantly, as shown in Figure \ref{motivation}(a), in a text-image retrieval task, an image is annotated by a unique text description, but this pattern seems to be inapplicable in the SBIR task. In all datasets employed for SBIR, images and sketches do not correspond one by one. This phenomenon exacerbates the domain gap, which impairs the compactness within the class and brings difficulties to constructing the metric space. To address this issue, most of prior works rigidly align multimodal feature representations belonging to the same class in the common metric space or align feature representations with additional modules. They narrow the domain gap by constructing image-sketch pairs as shown in Figure \ref{motivation}(b) or by constructing triples as shown in \ref{motivation}(c). However, all of them ignore the unpaired information between images and sketches and are unaware that how to utilize the non-one-by-one sketch-image pairs to eliminate or reduce the cross-domain gap is critical for SBIR.



To overcome the issues mentioned above, we propose a novel Three-Stream Joint Network (3JOIN) for the ZS-SBIR task, which incorporates image, edge map, and sketch into one unified end-to-end framework.
As shown in Figure \ref{motivation}(d), we employ the edge map as a bridge between the natural image and the sketch for the first time to minimize the
domain differences. Specifically, we exploit the one-to-one correspondence between natural images and edge maps to align natural images and edge map modalities. And we construct triples from the edge map and the sketch modalities to maintain the original similarity of the sketches in the retrieval space by metric learning.
Besides, our proposed method avoids the usage of side information. To resolve the semantic inconsistency between seen and unseen classes, we learns implicit semantics and maintains intra-class compactness without embedding the real semantics, achieving the current state-of-the-art retrieval performance.
The main contributions of 3JOIN are summarized as follows,

\begin{itemize}
\item \textbf{Brave idea:} A novel ZS-SBIR method called Three-Stream Joint Network (3JOIN) is proposed which possesses a 3-stream network to conduct joint training of images, sketches, and extracted edge maps. As far as we know, 3JOIN is the first one to introduce edge maps into the ZS-SBIR task.
\item \textbf{Novelty:} We design a new modality alignment strategy that treats edge maps as a bridge between natural images and sketches to narrow the domain gap between natural images and sketches.
\item \textbf{Scalability:} Our proposed method avoids the use of side information and maintains intra-class compactness without embedding real semantics, thus eliminating the need for advance knowledge of class names and reducing resource consumption for extracting real semantics.
\item \textbf{Technical quality:} Extensive experiments are conducted over two widely-used benchmark datasets, which demonstrate the superiority of our proposed method over several state-of-the-art baselines.
\end{itemize}

%
%
%
%

\section{Related Work}\label{sec:section_02}
In this section, we provide a brief overview of some recent literature in the field of SBIR, ZSL, and ZS-SBIR.

\subsection{Sketch-Based Image Retrieval}
The primary task of SBIR is to embed sketches and images into a common feature space while reducing the domain gap between sketch and image \cite{DengXWYT20, DeyR0LS19}. Generally, two strategies are proposed to narrow the domain gap. The first strategy extracts edge maps from natural images and the second one builds a common learning framework for sketches and natural images to learn domain transfer features in an end-to-end manner \cite{YuLSXHL16, SongYSXH17, GuoLWLWL17, Pang0YZHXS19, LinFLGXJ19}. Most pioneering SBIR methods utilize the first strategy, and some edge detection methods are used to convert natural images into edge maps. The geometric features are then extracted from the edge maps and sketches using well-designed descriptors, e.g., HOG descriptor \cite{HuC13}, histogram of oriented edges \cite{Saavedra14}, and Learned Key Shapes (LKS) \cite{SaavedraB15}. However, the direct use of geometric features to represent the samples inevitably loses some information and results in suboptimal results. In recent years, with the advances of deep neural networks, a large number of works have emerged to extract deep features from images and sketches using CNNs to learn better representations \cite{XuASRS18, BhuniaYHXS20, SainBYXS21, XuS00GZL21}. A representative work proposed by Qi et al. \cite{QiSZL16} uses a Siamese network to aggregate sketch and image features of the same category and separate features of different categories. Sangkloy et al. \cite{SangkloyBHH16} utilized a ranking loss function to constrain the feature distance between sketch and positive natural images to be smaller than the feature distance between sketch and negative natural images. And A further performance improvement is achieved by learning the cross-domain mapping through a pre-training strategy. By combining hashing framework and deep learning, Liu et al. \cite{LiuSSLS17} proposed a method that captured cross-domain similarity to improve retrieval performance. However, these methods utilize non-one-to-one sketch-image pairs or triples to rigidly align sketch and image domains, which all neglect unpaired information between images and sketches. In contrast, our proposed method builds one-to-one image and edge map pairs by extracting edge maps of natural images and uses edge maps as a bridge between the sketch domain and image domain. Besides, a three-stream joint training network explores the relationships among images, sketches, and edge maps.

\begin{figure*}
\begin{minipage}{0.245\linewidth}\centering
\centerline{\includegraphics[height=7.8cm]{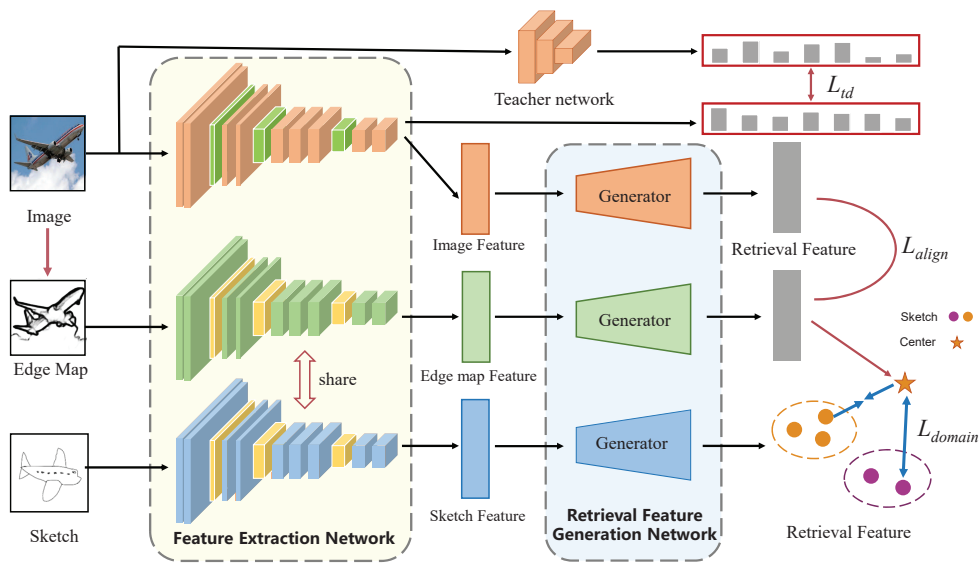}}
\end{minipage}\vspace{-0.1cm}
\caption{The framework of our proposed method. we convert natural images
into the form of edge maps and perform a three-stream joint training
network to explore the relationships among images, sketches, and edge maps. In the retrieval space, we use alignment loss and triplet loss to guide the joint training of the three branches.}\label{framework}
\end{figure*}

\subsection{Zero-Shot Learning}
Zero-Shot Learning (ZSL) expects models to have the ability of identifying new classes \cite{LiZZH18, MinYXWZZ20}. Existing zero-shot methods can be roughly divided into two categories, i.e., embedding-based methods and generation-based ones. Most embedding-based methods learn nonlinear multimodal embeddings \cite{XianLSA19, WeiY0DL19, Xie0J0ZQY019}. For example, SAE \cite{KodirovXG17} learns an auto-encoder that maps visual features to semantic embeddings and performs a nearest neighbor search in a semantic space. The generation-based methods utilize generators to synthesize features of unseen classes \cite{LongLSSDH17, WangPVFZCRC18, XianLSA18}. For instance, SP-AEN \cite{ChenZ00C18} preserves semantic information during image feature synthesis to ensure that the synthesized features carry more knowledge. It is worth noting that most ZSL methods use side information to transfer information from seen to unseen classes, such as text-based embedding or hierarchical embedding. Such a strategy would require prior knowledge of the exact class name and additional extraction work, which would result in a burden of resources. Besides, another popular trend is to use attributes as side information, which requires expensive annotations from professionals. By contrast, in this paper, we avoid employing side information to achieve good results under the ZSL scenario. By using a three-stream joint training network and teacher network, we successfully capture the underlying implicit properties of seen classes and transfer them to new classes.

\subsection{Zero-Shot Sketch-Based Image Retrieval}
Combining SBIR with the setting of ZSL, ZS-SBIR not only has to minimize the domain gap between images and sketches but also needs to transfer the knowledge gained in the seen classes to the unseen classes. ZS-SBIR task is facing huge challenges, but it has good prospects for real applications. The work \cite{Shen0S018} first combines ZSL with SBIR and proposes a cross-modal hashing method to mitigate the heterogeneity between two different modalities. CVAE \cite{YelamarthiRMM18} proposes a generative model based on both adversarial auto-encoder and variational auto-encoder, which takes sketch as input and generates additional details. SEM-PCYC \cite{0001A19} proposes a semantically tied paired cycle consistency generation model that maps the visual information of sketches and images to a common semantic space by adversarial training. SAKE \cite{0017XWY19} fine-tunes the pre-trained model to retain previously acquired knowledge during the teacher-student optimization process. Without using semantic embeddings, RPKD \cite{TianXWSL21} proposes knowledge distillation that maintains relationships to study generalizable embeddings. However, all of these methods ignore the aid of edge maps and the large domain differences introduced by the use of non-one-to-one sketch-image pairs. In this work, we utilize the alignment of edge maps to natural images to reduce the domain gap between images and edge maps and use prototypes of each category of edge maps to reduce the differences between sketches and edge maps.

\section{Our Method}

\subsection{Notations and Problem Definitions}
In ZS-SBIR task, a dataset is divided into two parts, the training set for building the model and the test set for evaluating the performance. Let $\mathcal{D}_{tr} = \{\mathcal{X}^{seen}, \mathcal{S}^{seen}\}$ denote the training dataset which contains $N_x$ images $\mathcal{X}^{seen} = \{x_i, y_i^x\}^{N_x}_{i=1}$ and $N_s$ sketches $\mathcal{S}^{seen} = \{s_i, y_i^s\}^{N_s}_{i=1}$, where $y_i^x, y_i^s \in \mathcal{C}^{seen}$ and $\mathcal{C}^{seen}$ represents the set of training classes. In contrast to the existing literature, the assumption of image-sketch pairing is not required in our method. Correspondingly, the testing set, denoted as $\mathcal{D}_{te} = \{\mathcal{X}^{unseen}, \mathcal{S}^{unseen}\}$, is unseen during training process and $\mathcal{X}^{unseen}=\{x_j, y_j^x\}^{M_x}_{j=1}, \mathcal{S}^{unseen}=\{s_j, y_j^s\}^{M_s}_{j=1}$, where $M_x$ and $M_s$ are the number of images and sketches in the test set, respectively, $y_j^x, y_j^s \in \mathcal{C}^{unseen}$, and $\mathcal{C}^{unseen}$ represents the set of testing classes. Under zero-shot setting, the sketch and image data of the seen classes are used for training only, that is, the categories of the training and testing set are disjoint, i.e., $\mathcal{C}^{seen} \cap \mathcal{C}^{unseen} = \varnothing$.
Given a specific sketch $s$ from $\mathcal{D}_{te}$, we aim to retrieve the corresponding images, e.g., with the same category as $s$ has, from the natural image gallery $\mathcal{X}^{unseen}$.

\subsection{Overall Architecture}
\subsubsection{Image Expansion}

As mentioned before, there are huge domain differences between sketches and natural images: different content and different styles. Natural images have multiple objects with rich colors and detailed textures, while sketches contain only the outline of an object. To minimize the differences between the two domains, we turn to the edge map for help. Being composed of black lines and white background, edge map of one instance has the same content information as to its corresponding natural image and a similar style to the sketch, which can serve as a bridge between image and sketch domains.

There are plenty algorithms for extracting edge maps, e.g., Canny detector \cite{Canny86a}, Gb \cite{LeordeanuSS12}, Fast Edge Detection Using Structured Forests \cite{DollarZ15}, and Bi-Directional Cascade Network (BDCN) \cite{HeZYSH19}. In a sense, the stronger the adopted edge extraction algorithm is, the more effective it is in painting the content of the natural image with black lines, and thus the more it can compensate for the domain differences between the image and the sketch. But what edge extraction algorithm is employed is not the main concern of our method. The sensitivity of our method to the edge map extraction algorithm will be analyzed in the subsequent experiment section. In this paper, we use BDCN \cite{HeZYSH19} as the edge map extraction method. We let $\mathcal{E} = \{e_i\}^{N_x}_{i=1}$ represent the edge maps of $\mathcal{X}^{seen}$, where $e_i$ is the extracted edge map of the $i$-th natural image $x_i$.

\subsubsection{Three-Stream Joint Network}
As shown in Figure 2, our method consists of three branches, i.e., the natural image branch, the edge map branch, and the sketch branch. The images, sketches, and extracted edge maps are fed into corresponding feature extraction networks and retrieval feature generation networks, respectively. In the retrieval space, we use alignment loss and triplet loss to explore the relationship among natural images, sketches and edge maps and guide the joint training of the three branches.

Specifically, each of branches employs a feature extraction network combining a series of convolutional layers and pooling layers to extract deep features. Mathematically, let $\mathcal{G}_x$, $\mathcal{G}_e$, and $\mathcal{G}_s$ denote the feature extraction network for natural images, edge maps, and sketches, respectively. Then, the features of the natural image are represented as $f^x = \mathcal{G}_x(\mathcal{X};\theta_x)$, the features of the edge map are $f^e = \mathcal{G}_e(\mathcal{E};\theta_e)$, and $f^s = \mathcal{G}_s(\mathcal{S};\theta_s)$ are the features of the sketch correspondingly. As mentioned before, the edge map and the sketch have similar styles, thus to ensure that the extracted feature points of them are similar, we set the feature extraction network weights of the edge map and the sketch to be shared, i.e., $\theta_e = \theta_s$. Compared with sketches and edge maps, natural images are usually more complex, which contain rich colors and detailed textures, so the feature extraction network weights of the image are not shared with the other two.

\subsection{Three-Stream Joint Training}
We extract edge maps from natural images and use corresponding edge maps as a bridge between natural images and sketches to minimize the domain gap between images and sketches. More specifically, we use alignment loss to align images and edge map modalities and triplet loss to increase the intra-class compactness of sketch and edge map modalities.

\subsubsection{Knowledge Distillation Loss}
We introduce the teacher network to resolve the semantic inconsistency between seen and unseen classes. Several works have demonstrated that the output of a teacher network can provide a fine-grained semantic pseudo-label to a student network \cite{HintonVD15, FurlanelloLTIA18}, because the teacher network usually is equipped with a powerful feature mapping function trained on a large-scale dataset (e.g., ImageNet \cite{DengDSLL009}). Introducing the teacher network into ZS-SBIR has two advantages: 1) The image feature extraction network as a student network mimics the reaction of the teacher network, which facilitates better feature extraction and fast convergence of the student network. 2) The weights of the teacher network are not updated, so the pseudo-labels it provides to the student network have stable implicit semantics. Image features are learned under the guidance of implicit semantics, which facilitates the student network to transfer the learned knowledge to unseen classes. Specifically, a natural image is fed into the teacher network, a classifier that has been trained on $C$ classes, to obtain a prediction $\mathbf{q}_i = $ Softmax($\mathbf{t}_i$) $\in \mathbb{R}^C$, where $\mathbf{t}_i$ is the embeddings of the $i$-th natural image in the teacher network. We expect the student network to make the same prediction for the same sample, which is denoted as $\mathbf{p}_i = $ Softmax($\mathbf{x}_i$) $\in \mathbb{R}^C$. Besides, to achieve inter-modal migration, we add a parameter divergence loss, which is as follows,
\begin{equation}
L_{kd }=\frac{1}{N} \sum_{i=1}^{N} \mathcal{K}\left(\mathbf{q}_i, \mathbf{p}_i\right) + \gamma \left\|\theta_x-\theta_s\right\|_{F}^{2},
\end{equation}
where $\gamma$ is a parameter, $N$ is the number of samples in a mini-batch, and $\mathcal{K}(\cdot\ ,\ \cdot)$ measures differences between two probability distributions, which is defined as follows,
\begin{equation}
\mathcal{K}(\mathbf{p}, \mathbf{q})=\sum_{i}^N-\mathbf{q}_i \log \left(\mathbf{p}_i\right).
\end{equation}

\subsubsection{Alignment Loss}
Considering the one-by-one correspondence between an image and its corresponding edge map, alignment loss is proposed. In the retrieval space, even though the image and its corresponding edge map are from different modalities, their retrieval features should be the same because they are essentially different styles of the same sample. Therefore, to forcibly align the retrieved features of the image and edge map, the alignment loss function is defined as follows,
\begin{equation}
L_ {align}=\left\|\mathcal{F}_x(f^x;\psi_x)-\mathcal{F}_e(f^e;\psi_e)\right\|_{2}^{2},
\end{equation}
where $\mathcal{F}_x$ and $\mathcal{F}_e$ denote the retrieval feature generation network of natural images parameterized by $\psi_x$ and the retrieval feature generation network of edge maps parameterized by $\psi_e$, respectively. By introducing alignment loss, we reduce the distance between the one-to-one image and edge map retrieved features, giving a new way to narrow the domain gap between images and edge maps. Such a strategy also contributes to shrinking the domain gap between images and sketches.

\subsubsection{Sketch-Edge Loss}
Alignment loss cannot be used directly between sketches and edge maps, because sketches and edge maps do not correspond one-to-one, e.g., sketches are created from user scribbles while edge maps are extracted from natural images. And compared to sketches, there is more variation within the edge map class since it contains background information and other objects. Therefore, we generate a center for each class of the edge map, and the alignment loss between sketch and class center of edge map modality is defined as follows,
\begin{equation}
L_{center}=\frac{1}{N} \sum_{i=1}^{N}\left\|\mathcal{F}_s(f^s_i;\psi_s)-c_y\right\|_{2}^{2},
\end{equation}
where $\mathcal{F}_s$ is the retrieval feature generation network of sketches  parameterized by $\psi_s$, $f^s_i$ is the feature of the $i$-th sketch labeled $y$, and $c_y$ is the center of class $y$ of the edge map modality. By aligning the retrieved features of the sketch with the class centers of the edge map, intra-class variation in the sketch can be minimized and the relationship between the sketch and the edge diagram can be constrained to some extent.

Further considering the relationship between sketches and edge maps, sketches and edge maps can be treated as the same modality data, since sharing a similar style. Therefore, in the retrieval space, we expect that the distance between similar sketch and edge map pairs should be smaller than that between negative pairs. Specifically, given a sketch $f^s_i$ belonging to class $y$, let the corresponding class center of the edge map, i.e., $c_y$ be the anchor, we hope:
\begin{equation}
\mathcal{R}\left(c_y, \mathcal{F}_s(f^s_i;\psi_s)\right)<\mathcal{R}\left(c_y, \mathcal{F}_s(f^s_j;\psi_s)\right), y=y_{i} \neq y_{j},
\end{equation}
where $\mathcal{R}(\cdot\ ,\ \cdot)$ measures Euclidean distance. The reason that the class center of the edge map is the anchor point is that the edge map is the bridge between the image and the sketch, and the natural image and the edge map are already aligned in the retrieval space. So, the distance relationship between positive and negative sketches and edge maps can be migrated to positive and negative sketches and the images. And the Sketch-Edge triplet loss is defined as,
\begin{equation}
\begin{aligned}
\mathcal{L}_{triplet }=\frac{1}{N} \sum_{i=1}^{N}[\max \{0, \mu&+\mathcal{R}\left(c_y, \mathcal{F}_s(f^s_i;\psi_s)\right) \\
&-\mathcal{R}\left(c_y, \mathcal{F}_s(f^s_j;\psi_s)\right)\}],
\end{aligned}
\end{equation}
where $\mu$ is a hyper-parameter denoting the margin. For each sample in a mini-batch, we select the hardest negative sketch, which is the closest sample to the anchor point among those that are not in the same class as the anchor point and its retrieval feature is denoted as $\mathcal{F}_s(f^s_j;\psi_s)$.

The overall Sketch-Edge loss is given as follows,
\begin{equation}
\begin{aligned}
\mathcal{L}_{domain }= \mathcal{L}_{center } + \eta \mathcal{L}_{triplet },
\end{aligned}
\end{equation}
where $\eta$ is a balance parameter.

It is worth noting that in practice it is not realistic to calculate the class center of all data due to the huge waste of resources. In this paper, we utilize the centers of each mini-batch instead of the centers of all data. Due to the small size of the mini-batch, to avoid data perturbation, we update the centers of each mini-batch cumulatively with the following equation,
\begin{equation}
c_{y}^{t}=\frac{c_y^{t-1} * n^{t-1} + \sum_{i=1}^{N} \mathcal{I}\left(y_{i}=y\right)\mathcal{F}_e(f^e_i;\psi_e)}{n^{t-1}+\sum_{i=1}^{N} \mathcal{I}\left(y_{i}=y\right)},
\end{equation}
where $c_y^{t-1}$ is the centers of class $y$ in round $t-1$, $n^{t-1}$ is the number of samples belonging to class $y$ before round $t-1$, $N$ is the size of mini-batch, and $\mathcal{I}$ is an instruction function, where $\mathcal{I}\left(y_{i}=y\right)=1$ when $y_{i}=y$ and $\mathcal{I}\left(y_{i}=y\right)=0$ otherwise.

\subsubsection{Classification Loss}
We propose a new scheme to efficiently constrain the relationship among images, edge maps, and sketches and implement joint training through alignment loss and Sketch-Edge loss. However, both of alignment loss and Sketch-Edge loss fail to focus on the intra-class compactness. Images (sketches) in the same category should have similar retrieval features. Based on this, to learn the discriminative representations, we introduce a classifier and use a cross-entropy loss to align the learned features with their labels, which is defined as follows,
\begin{equation}
\mathcal{L}_{cls}=-\sum_{i=1}^{N} \log \frac{\exp \left(\alpha_i^{\top} \mathcal{F}_m(f^m_i;\psi_m)+\beta_i\right)}{\sum_{j \in \mathcal{C}^{\text {seen }}} \exp \left(\alpha_{j}^{\top} \mathcal{F}_m(f^m_i;\psi_m)+\beta_{j}\right)},
\end{equation}
where $\alpha$ and $\beta$ are the weight and bias of the classifier separately, and $m \in \{image, sketch\}$.

\subsection{Overall Objective Function}
Jointly considering the losses defined above, i.e., knowledge distillation loss, alignment loss, Sketch-Edge loss, and classification loss, we define the full objective of our model as follows,
\begin{equation}
\mathcal{L}=\mathcal{L}_{kd}+\lambda_{1} \mathcal{L}_{align}+\lambda_{2} \mathcal{L}_{domain} + \lambda_{3} \mathcal{L}_{cls},
\end{equation}
where $\lambda_{1}$, $\lambda_{2}$, and $\lambda_{3}$ are the trade-off parameters.

\section{Experiment}
In this section, we first introduce the datasets and the experimental settings including implementation details and evaluation protocol. And then we provide the experimental results compared with the State-of-the-Art and further analysis.

\subsection{Datasets}
We conducted extensive experiments to evaluate the performance of our proposed method on two widely-used  large-scale datasets, i.e., Sketchy \cite{SangkloyBHH16} and TU-Berlin \cite{EitzHA12}. Both datasets consist of data from two modalities, i.e., images and sketches.



$\textbf{Sketchy}$ is a large-scale dataset containing $125$ categories. Initially, there are $100$ images and at least $600$ sketches for each category. In \cite{LiuSSLS17}, an extended version is proposed which has additional  $60,502$ natural images collected from ImageNet. Thus, a total of $73,002$ natural images and $75,479$ sketches are available from $125$ categories. Following the setting in \cite{0017XWY19}, $100$ categories are used for training and other $25$ categories form the test set.

$\textbf{TU-Berlin}$ contains a total of $20,000$  free-hand sketches labeled by $250$ categories. Liu et al. \cite{LiuSSLS17} collected $204,489$ images to extend this dataset so as to adapt it for SBIR task. For a fair comparison
with state-of-the-art methods, we adopted the same data split setting as \cite{0017XWY19}. We randomly selected $30$ categories for testing and the remaining $220$ categories are left for training. It is worth noting that each testing category requires at least $400$ natural images to satisfy the retrieval.

\subsection{Experimental Settings}

\subsubsection{Implementation Details}
All experiments are performed under the environment of GTX 1080 Ti GPU. We implemented 3JOIN on PyTorch with the Adam optimizer with $\beta_1 = 0.9, \beta_2 = 0.999$. The initial learning rate is set to $1 \times 10^{-4}$. The batch size and the number of maximum training epochs are set as $24$ and $10$, respectively. We set $\gamma, \lambda_1$, $\lambda_2$, $\lambda_3$, and $\eta$ to $\{100, 0.1, 0.1, 0.1, 10\}$ on Sketchy and $\{100, 0.01, 0.01, 1, 100\}$ on TU-Berlin. The margin in triplet loss, i.e., $\mu$, is set to $0.2$ on all  datasets. In our proposed method, the feature extraction network for each branch and the teacher network for knowledge distillation are constructed based on ResNet-50 \cite{HeZRS16}. The parameters of the teacher network are frozen throughout the training process. In addition, the Bi-Directional Cascade Network \cite{HeZYSH19} is applied to extract the edge map of each natural image during the training stage.

\subsubsection{Evaluation Protocol}
We conducted a cross-modal retrieval task to evaluate the performance of 3JOIN, i.e., retrieving similar natural images given a sketch as a query, and the cosine distance is used in sorting the retrieval results. In this paper, two widely-used evaluation criteria are adopted, i.e., Mean Average Precision (mAP) and Precision (Prec). For these metrics, a larger value indicates better performance. For a fair comparison with previous hashing based works, we encoded the real-valued retrieval features as binary codes, so that the retrieval speed can be significantly improved. Specifically, we employed the iterative quantization (ITQ) algorithm \cite{GongLGP13} to generate hash codes based on real-valued retrieval features and calculated the Hamming distance to sort the retrieval results in Hamming space.

\begin{table}[t]\normalsize \center
\caption{Overall comparison of our proposed method with other methods using semantic embedding. "†" indicates the retrieval features obtained by hash codes  in the original paper. The best and sub-best results are bolded and underlined, respectively.} \vspace{-0.2cm}
\setlength{\tabcolsep}{2.1mm}{
\begin{tabular}{c c cc cc}\toprule[1pt]
\multirow{3}{*}{Methods}&\multirow{3}{*}{ Dim}&\multicolumn{2}{c}{Sketchy} &\multicolumn{2}{c}{TU-Berlin} \\\cmidrule(lr){3-4}
\cmidrule(lr){5-6} & &mAP&Prec&mAP&Prec\\
 & &@all&@100&@all&@100\\
\hline
 {SAE \cite{KodirovXG17}}&{300}&{0.210}&{0.302}&{0.161}&{0.210}\\
 {ZSH\cite{0002LCSSS16}}&{$64^\dagger$}&{0.165}&{0.217}&{0.139}&{0.174}\\
 {ZSIH \cite{Shen0S018}}&{$64^\dagger$}&{0.254}&{0.340}&{0.220}&{0.291}\\
 {SEM-PCYC \cite{0001A19}}&{$64^\dagger$}&{0.344}&{0.399}&{0.293}&{0.392}\\
 {SEM-PCYC \cite{0001A19}}&{$64$}&{0.349}&{0.463}&{0.297}&{0.426}\\
 {SAKE \cite{0017XWY19}}&{$64^\dagger$}&\underline{0.364}&\underline{0.487}&\underline{0.359}&\underline{0.481}\\
 {3JOIN}&{$64^\dagger$}&\textbf{0.462}&\textbf{0.595}&\textbf{0.361}&\textbf{0.487}\\
 \hdashline
 {SAKE \cite{0017XWY19}}&{$512$}&{0.547}&{0.692}&\underline{0.475}&\underline{0.599}\\
 {SAKE+AMDReg \cite{DuttaSB20}}&{$512$}&\underline{0.551}&\underline{0.715}&{0.447}&{0.574}\\
 {OCEAN \cite{ZhuXSLWS20}}&{$512$}&{0.462}&{0.590}&{0.333}&{0.467}\\
 {3JOIN}&{$512$}&\textbf{0.620}&\textbf{0.724}&\textbf{0.496}&\textbf{0.613}\\
 \hline
\toprule[1pt]\end{tabular}} \label{map_yes}\end{table}

\subsection{Comparison with the State-of-the-Art}
To verify the superiority of 3JOIN, we compared our proposed method with several existing state-of-the-art approaches, including two representative works of ZSL (i.e., SAE \cite{KodirovXG17} and ZSH \cite{0002LCSSS16}), three prior works on SBIR (i.e., GN Triplet \cite{SangkloyBHH16}, Siamese CNN \cite{QiSZL16}, and DSH \cite{LiuSSLS17}), and nine existing works on ZS-SBIR (i.e., ZSIH \cite{Shen0S018}, SEM-PCYC \cite{0001A19}, SAKE \cite{0017XWY19}, SAKE+AMDReg \cite{DuttaSB20}, OCEAN \cite{ZhuXSLWS20}, CAAE \cite{YelamarthiRMM18}, DSN \cite{Wang0YWD21}, RPKD \cite{TianXWSL21}, and SBTKNet \cite{TURSUN2022108528}). Among them, SAE, ZSH, ZSIH, SEM-PCYC, SAKE, SAKE+AMDReg, and OCEAN utilize language models to extract side information, i.e., word model and hierarchical information to bridge the inconsistency between the seen and unseen classes, while the remaining methods and our 3JOIN do not utilize any side information.
The performance of our proposed 3JOIN and the state-of-the-art methods on Sketchy and TU-Berlin are shown in Table \ref{map_yes} and Table \ref{map_no}. To fully demonstrate the performance of our approach, we reported the results where the retrieval features are represented as 64-dimensional binary codes and the retrieval features are 512-dimensional real values.

The results of our proposed method and several works using semantic embedding are reported in Table \ref{map_yes}, including two ZSL methods and five ZS-SBIR methods. From this table, we can find:

\begin{itemize}[leftmargin=*]
\item The performances of ZSL methods fall far behind those of ZS-SBIR methods, probably because the ZSL methods fail to handle the inter-modal differences, making the embedding of sketches and images in different spaces without considering the inter-modal consistency.
\item Among these methods using side information, SAKE+AMDReg and SAKE obtain the best performance on TU-Berlin and Sketchy, respectively. AMDReg is a regularization method used seamlessly with the ZS-SBIR method to improve their performance. SAKE employs the teacher network to extract visual features from complex images as our method, however, it additionally uses side information for constructing a similarity matrix between labels to guide the teacher-student optimization process. This strategy requires the exact class name to be known in advance and additional extraction work. The results in Table \ref{map_yes} shows that 3JOIN achieves better performance than SAKE without using any semantic information. This is because our method successfully reduces the domain gap between natural images and sketches using 3-stream joint training with the help of edge maps.
\end{itemize}

\begin{table}[t]\normalsize \center
\caption{Overall comparison of 3JOIN with other methods without semantic embedding. "†" indicates the retrieval features obtained by hash codes, and "-" indicates that the corresponding results are not reported. The best and sub-best results are bolded and underlined, respectively.} \vspace{-0.2cm}
\setlength{\tabcolsep}{2.1mm}{
\begin{tabular}{c c cc cc}\toprule[1pt]
\multirow{3}{*}{Methods}&\multirow{3}{*}{ Dim}&\multicolumn{2}{c}{Sketchy} &\multicolumn{2}{c}{TU-Berlin} \\\cmidrule(lr){3-4}
\cmidrule(lr){5-6} & &mAP&Prec&mAP&Prec\\
 & &@all&@100&@all&@100\\
\hline
 {GN Triplet \cite{SangkloyBHH16}}&{1024}&{0.211}&{0.310}&{0.189}&{0.241}\\
 {CAAE \cite{YelamarthiRMM18}}&{4096}&{0.196}&{0.284}&{-}&{-}\\
 {Siamese CNN \cite{QiSZL16}}&{64}&{0.132}&{0.175}&{0.109}&{0.141}\\
 {DSH \cite{LiuSSLS17} }&{$64^\dagger$}&{0.164}&{0.210}&{0.122}&{0.198}\\
 {DSN \cite{Wang0YWD21} }&{$64^\dagger$}&\underline{0.436}&\underline{0.553}&\textbf{0.385}&\textbf{0.497}\\
 {RPKD \cite{TianXWSL21}}&{$64^\dagger$}&{0.423}&{0.536}&\underline{0.361}&\underline{0.491}\\
 {3JOIN}&{$64^\dagger$}&\textbf{0.462}&\textbf{0.595}&\underline{0.361}&{0.487}\\
 \hdashline
 {DSN \cite{Wang0YWD21}}&{512}&{0.583}&{0.704}&{0.481}&{0.586}\\
 {SBTKNet \cite{TURSUN2022108528}}&{512}&{0.553}&{0.698}&{0.480}&{0.608}\\
 {RPKD \cite{TianXWSL21}}&{512}&\underline{0.613}&\underline{0.723}&\underline{0.486}&\underline{0.612}\\
 {3JOIN}&{$512$}&\textbf{0.620}&\textbf{0.724}&\textbf{0.496}&\textbf{0.613}\\
 \hline
\toprule[1pt]\end{tabular}} \label{map_no}\end{table}


Besides, we listed the results of 3JOIN and several methods without semantic embedding in Table \ref{map_no}, e.g., three SBIR methods and four ZS-SBIR methods.
From these results, we can draw the following observations:
\begin{itemize}[leftmargin=*]
\item Except for CAAE, the performances of ZS-SBIR methods are significantly better than those of the SBIR methods. One possible reason is that none of the SBIR methods are specifically designed for retrieving unseen categories, and none of them consider how to transfer the knowledge learned in the seen labels to the unseen ones; therefore, most of them overfit in the training set and lose the ability to generalize to new categories.
\item Among these methods, Siamese CNN is the only method using edge map, but it is an SBIR method that leaves out the performance on new categories. Besides, it only considers the relationship between sketch and edge map captured with a siamese network, and fails to catch the relationship between image and edge map, which causes its performance to be much lower than our method.
\item The early work, i.e., CAAE uses a relatively simple framework with a single autoencoder or reconstruction loss to capture the relationships between images and sketches. More recent works such as DSN, RPKD, and SBTKNet use more complex structures and achieve better performance, but still cannot surpass our method in most cases.
\item 3JOIN achieves mAP value of $0.620$ and $0.494$ on Sketchy and TU-Berlin for 512-dimensional retrieval features, which demonstrates the effectiveness of exploiting a sufficient combination of sketches, natural images, and edge maps for addressing cross-domain ZS-SBIR problems.
\end{itemize}

\begin{table}[t]\normalsize \center
\caption{Ablation results of individual component in 3JOIN with 512 dimensional retrieval features on the Sketchy and TU-Berlin datasets. The best results are bolded.} \vspace{-0.1cm}
\setlength{\tabcolsep}{1.5mm}{
\begin{tabular}{l cc cc}\toprule[1pt]
\multirow{3}{*}{Variants}&\multicolumn{2}{c}{Sketchy} &\multicolumn{2}{c}{TU-Berlin} \\\cmidrule(lr){2-3}
\cmidrule(lr){4-5} &mAP&Prec&mAP&Prec\\
&@all&@100&@all&@100\\
\hline
 {1. baseline($\mathcal{L}_{cls}$)}&{0.278}&{0.362}&{0.316}&{0.426}\\
 {2. $\mathcal{L}_{cls} + \mathcal{L}_{kd}$}&{0.367}&{0.460}&{0.343}&{0.442}\\
 {3. $\mathcal{L}_{cls}+\mathcal{L}_{kd}+\mathcal{L}_{align} $}&{0.593}&{0.701}&{0.488}&{0.601}\\
 {4. $\mathcal{L}_{cls}+\mathcal{L}_{kd}+\mathcal{L}_{domain} $}&{0.591}&{0.707}&\textbf{0.497}&{0.612}\\
 {5. $\mathcal{L}_{cls}+\mathcal{L}_{kd}+\mathcal{L}_{align} + \mathcal{L}_{center} $}&{0.593}&{0.702}&{0.494}&{0.613}\\
 {6. $\mathcal{L}_{cls}+\mathcal{L}_{kd}+\mathcal{L}_{align} + \mathcal{L}_{triplet} $}&{0.609}&{0.729}&{0.495}&{0.611}\\
 {7. 3JOIN}&\textbf{0.620}&\textbf{0.724}&{0.496}&\textbf{0.613}\\
 \hline
\toprule[1pt]\end{tabular}} \label{ablation}\end{table}

\subsection{Further Analysis}
\subsubsection{Ablation Experiments}
To analyze the impact of key components of 3JOIN, i.e., knowledge distillation, alignment, Sketch-Edge domain gap, and classification, we ablated their corresponding loss terms and the results conducted on Sketchy and TU-Berlin are reported in Table \ref{ablation}. For the convenience of presentation, we index all variants and our proposed method from top to bottom in Table \ref{ablation} as model 1-7.

\begin{figure*}
\begin{minipage}{0.245\linewidth}\centering
\centerline{\includegraphics[height=6.3cm]{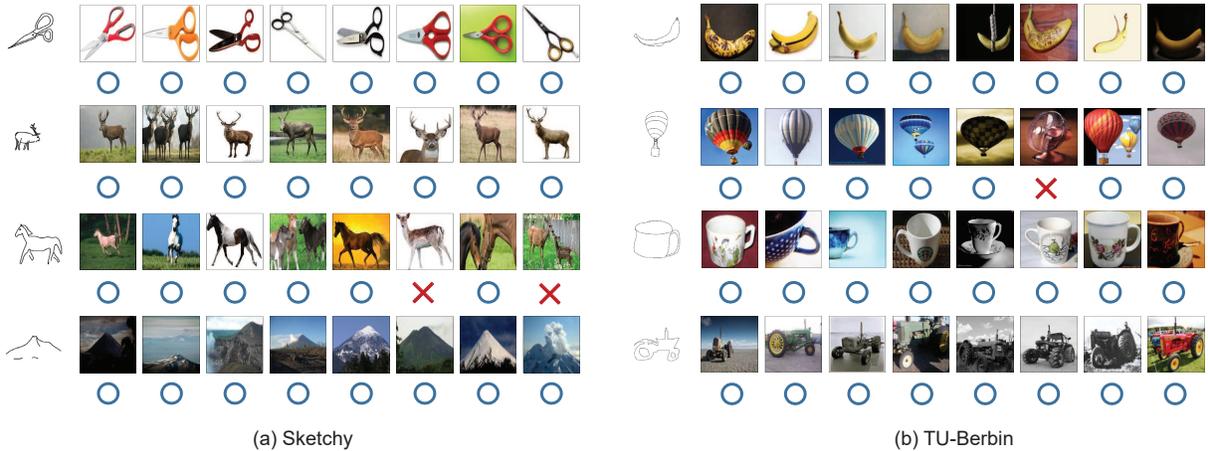}}
\end{minipage}
\caption{Top-8 retrieval results of testing samples on Sketchy and TU-Berlin with 512 dimensional retrieval features. The blue circles stand for correctly retrieved candidates while the red crosses indicate wrong retrieved candidates .}\label{visual}
\end{figure*}

To verify the validity of each component, we first trained a baseline which maps sketches and natural images into a common embedding space with only classification loss. From this table, we have the following observations:
\begin{itemize}[leftmargin=*]
\item Adding knowledge distillation loss to the baseline, the performance has been improved significantly, for example, the MAP value on Sketchy is increased from $0.278$ to $0.367$. A possible reason is that the image feature extraction network mimics the responses of the teacher network facilitating the student network to better extract the image feature extraction and transfer the learned implicit semantics to the unseen categories.
\item Model 3 outperforms Model 2, demonstrating that by explicitly aligning the natural image and the corresponding edge map, better retrieval features can be obtained.
\item Model 4 adds Sketchy-Edge domain loss and performs better than Model 2 in all cases, revealing the importance of considering the relationship between sketches and edge maps.
\item Compared with model 3, both models 5 and 6 achieve better results indicating the effectiveness of center loss and triplet loss and the fact that  encouraging retrieval features to maintain the similarity of samples in the original space through metric learning is beneficial for the ZS-SBIR task.
\item The complete model containing all learning losses and network components outperforms the rest of all variants in most cases.
\end{itemize}

\begin{table}[t]\normalsize \center
\caption{Comparisons among different edge map extraction algorithm with 512 dimensional retrieval features on the Sketchy and TU-Berlin datasets. The best results are bolded.}
\setlength{\tabcolsep}{3.6mm}{
\begin{tabular}{c cc cc}\toprule[1pt]
\multirow{3}{*}{Methods}&\multicolumn{2}{c}{Sketchy} &\multicolumn{2}{c}{TU-Berlin}\\ \cmidrule(lr){2-3} \cmidrule(lr){4-5} &mAP&Prec&mAP&Prec\\
&@all&@100&@all&@100\\
\hline
 {Canny \cite{Canny86a}}&{0.594}&{0.702}&{0.471}&{0.581}\\
 {Gb \cite{LeordeanuSS12}}&{0.609}&{0.718}&{0.485}&{0.599}\\
 {FEDSF \cite{DollarZ15}}&{0.601}&{0.709}&{0.486}&{0.605}\\
 {BDCN \cite{HeZYSH19}}&\textbf{0.620}&\textbf{0.724}&\textbf{0.496}&\textbf{0.613}\\
 \hline
\toprule[1pt]\end{tabular}} \label{edge} \vspace{-0.1cm} \end{table}

\subsubsection{Effect of Edge Map Extraction Algorithm}
To analyze the influence of the edge map extraction algorithm, we conducted experiments on Sketchy and TU-Berlin and the results are shown in Table \ref{edge}. Canny is a classical edge extraction algorithm proposed in 1986, which uses Gaussian smoothing filtering to remove noise, non-maximum suppression to remove pixels with insufficient gradient, and a double-thresholding algorithm to detect and connect edges. However, it achieves inferior performance due to its early and simple model. Both Gb and Fast Edge Detection using Structured Forests (FEDSF)  algorithms use geometric features to detect edges while BDCN extracts a multi-scale representation of the image using CNN and monitors each layer individually using a bidirectional pseudo-cascade structure to achieve layer-specific edge detection. From this table, we can find that BDCN outperforms Gb and FEDSF. In general, the stronger the adopted edge extraction algorithm is, the more effective it is in painting the content of the natural image with black lines, and thus the more it can compensate for the domain differences between the image and the sketch. According to the experimental results, in this paper, we adopted BDCN \cite{HeZYSH19} as the edge map extraction algorithm.

\subsubsection{Qualitative results}
To thoroughly evaluate the effectiveness of 3JOIN, we provided the top 8 natural image results in Figure \ref{visual} when a sketch from the unseen class comes as a query. Blue circles indicate correctly retrieved candidates, while red crosses stand for incorrectly retrieved samples. From this figure, we can observe that most of the retrieved candidates are in the same category of their query. However, the sixth retrieval result for querying, i.e., hot air balloons, on the TU-Berlin dataset is incorrect, mainly probably due to the similarity of the shape of the fan and the hot air balloon. Since the sketch lacks visual cues such as color, texture, and background, the model is confused by the examples with similar shapes. In addition, we report a more difficult retrieval situation where the retrieval results of two similar categories, i.e., deers and horses, are provided on the Sketchy dataset. It can be seen that most of the candidates are correct, except for the sixth and eighth candidates for horse retrieval, which indicates that our method is capable of discriminating finer-grained knowledge. The possible reason is that our method successfully maintains intra-class compactness and inter-data similarity using a metric learning approach.

\section{Conclusion}
In this paper, we propose a novel Zero-Shot Sketch-Based Image Retrieval method, i.e., Three-Stream Joint Network (3JOIN for short). To the best of our knowledge, it is the first attempt to introduce edge maps into the ZS-SBIR task. The proposed model explores the relationships among images, sketches, and edge maps by constructing a three-stream joint training network. Specifically, we design a new modality alignment strategy that explicitly aligns the image and edge map modalities through alignment loss and maintains the original similarity of the samples in the retrieval space through sketch-edge loss. Besides, 3JOIN avoids the usage of side information, learns implicit semantics and maintains intra-class compactness without embedding the real semantics, thus eliminating the need for advance knowledge of class names and reducing resource consumption for extracting real semantics. Extensive experiments on two real-world benchmarks have been conducted and the results demonstrate the superiority of our proposed method over several state-of-the-art baselines.

\clearpage

\bibliographystyle{ACM-Reference-Format}
\balance
\bibliography{sample-base}

\end{document}